\documentclass{llncs}
\usepackage{subfig}
\usepackage{graphicx}
\usepackage{color}
\usepackage{amsmath}
\usepackage{amssymb}
\usepackage{xspace}
\usepackage{url}
\usepackage{hyperref}

\makeatletter
\DeclareRobustCommand\onedot{\futurelet\@let@token\@onedot}
\def\@onedot{\ifx\@let@token.\else.\null\fi\xspace}

\newcommand{\figcaption}[1]{\def\@captype{figure}\caption{#1}}
\newcommand{\tblcaption}[1]{\def\@captype{table}\caption{#1}}
\makeatother

\def\title#1{{\noindent\Large{\bf #1}\par}}
\def\author#1{\begin{center}{\sc #1\par}\end{center}}
\renewcommand{\section}[1]{\vspace{0.1in}\noindent{\large\bf{#1}}\par\vspace{.05in}\par\nopagebreak}

\makeatletter
\def\thebibliography#1{\section{Recommended Readings\@mkboth
{REFERENCES}{REFERENCES}}\list
{[\arabic{enumi}]}{\settowidth\labelwidth{[#1]}\leftmargin\labelwidth
\advance\leftmargin\labelsep
\usecounter{enumi}}
\def\newblock{\hskip .11em plus .33em minus .07em}
\sloppy\clubpenalty4000\widowpenalty4000
\sfcode`\.=1000\relax}
\makeatother

\begin{document}
\pagestyle{empty}

\title{View-invariant action recognition}
\author{Yogesh Rawat \footnote{corresponding author},\\CRCV, University of Central Florida, Orlando, Florida, USA.}
\author{Shruti Vyas, \\CRCV, University of Central Florida, Orlando, Florida, USA.}

\section{Synonyms}
\begin{itemize}
\item{Cross-view action recognition}
\item{View-invariant action classification}
\item{View-invariant activity recognition}
\end{itemize}

\section{Related Concepts}
\begin{itemize}
\item{View-invariance}
\item{Action recognition}
\item{Activity classification}
\end{itemize}


\section{Definition}
Recognizing human actions from previously seen viewpoints is relatively easy when compared with unseen viewpoints. View-invariant action recognition aims at recognizing human actions from unseen viewpoints.


\section{Background}



\begin{figure}[t]
\begin{center}
   \includegraphics[width=0.3\linewidth]{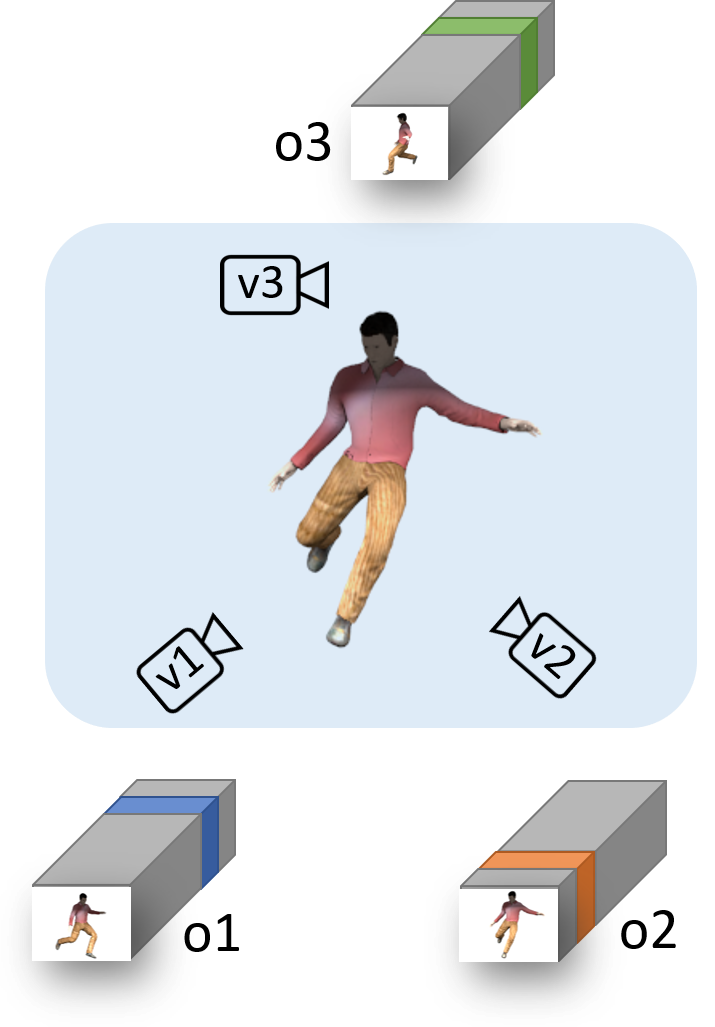}
\end{center}
\caption{{An action captured from different viewpoints (v1, v2, and v3) providing different observations (o1, o2, and o3) \cite{vyas2018time}.}}
\label{action_views}
\end{figure}

Human action recognition is an important problem in computer vision. It has a wide range of applications in surveillance, human-computer interaction, augmented reality, video indexing, and retrieval. The varying pattern of spatio-temporal appearance generated by human action is key for identifying the performed action. We have seen a lot of research exploring this dynamics of spatio-temporal appearance for learning a visual representation of human actions. However, most of the research in action recognition is focused on some common viewpoints \cite{duarte2018videocapsulenet}, and these approaches do not perform well when there is a change in viewpoint. Human actions are performed in a 3-dimensional environment and are projected to a 2-dimensional space when captured as a video from a given viewpoint. Therefore, an action will have a different spatio-temporal appearance from different viewpoints. As shown in Figure \ref{action_views}, observation o1 is different from observation o2 and so on. The research in view-invariant action recognition addresses this problem and focuses on recognizing human actions from unseen viewpoints.

\begin{figure}[t]
\begin{center}
   \includegraphics[width=0.9\linewidth]{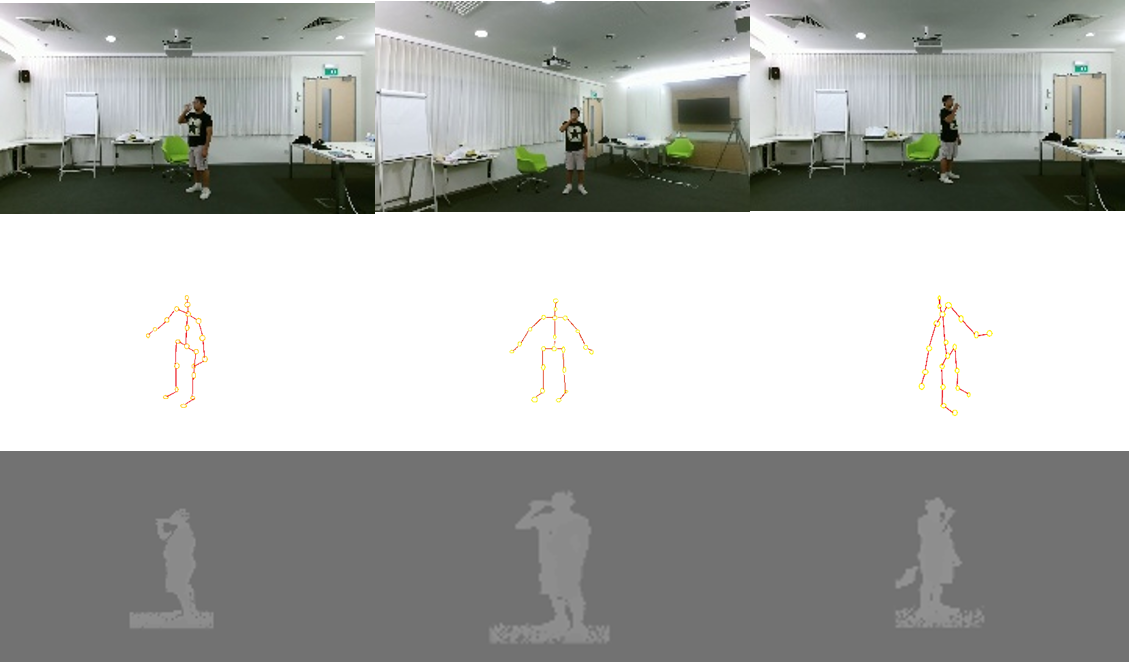}
\end{center}
\caption{{Video frames showing action in different modalities as seen from three different view-points \cite{shahroudy2016ntu}, Row-1: RGB, Row-2: Skeleton, and Row-3: Depth.}}
\label{modalities}
\end{figure}

There are different data modalities which can be used for view-invariant representation learning and perform action recognition. These include RGB videos, skeleton sequences, depth information, and optical flow. The skeleton sequences and depth information require additional sensors and are comparatively more difficult to capture when compared with RGB videos. Similarly, the optical flow is computationally expensive and require extra processing on RGB videos. These modalities can be used independently as well as in combination to solve the problem of view-invariant action recognition. Figure \ref{modalities} shows a sample instances activities performed by an actor which is captured from three different viewpoints in three different modalities.


Human action recognition from video sequences involves the extraction of visual features and encoding the performed action in some meaningful representation which can be used for interpretation. The view-invariant encoding of actions involves a lot of challenges, and there are different ways to address them. One possible solution is to track the motion as it evolves with the performed action. The track in itself will be invariant to any change in the viewpoint and can be used to extract a view-invariant representation of actions. Another approach is to analyze the spatio-temporal volume, which is covered by a human while performing any action. It is interesting to observe that such spatio-temporal volumes will have some similarities for same actions, and they can be useful to address change in viewpoints. The tracking of a human body can be useful to some extent, but human joints can move independently while performing most of the activities. Therefore, tracking of skeleton joints independently is very important for understanding human actions. The availability of large-scale training data has also enabled us to learn view-invariant representations using deep learning. We will cover the details of these approaches in the following sections.

\section{View-invariance in human actions}



The dynamics of body parts and the change in appearance play an important role in understanding human actions. These two properties can be effectively used to determine human actions in a video stream if there is no change in viewpoint. However, it becomes challenging when there is a change in the viewpoint, as the dynamics, as well as the appearance, will change with it significantly.  Therefore, it is important to represent human actions in such a way that the representation is invariant to any change in viewpoint. The idea is to encode the human action with a representation which does not change with change in viewpoint. We can observe in Figure \ref{modalities} how the appearance of the activity changes when seen from different viewpoints. It does not matter how the action is captured; this variation is present in all the modalities, including RGB, skeleton, and depth. 

\subsection{Motion trajectories}
An action sequence performed by any person will have a motion trajectory, which will be different for different actions. These motion trajectories can be useful in extracting rich view-invariant representation for action classification. The actions are performed in a 3D environment, and therefore, the corresponding motion trajectories are in 3D space. The change in speed and direction of the trajectory plays an important role in inferring the performed action. The continuities and discontinuities in position, velocity, and acceleration in a 3D trajectory are preserved in 2D trajectories under a continuous projection \cite{rao2002view}. Therefore, we can use the 2D spatio-temporal curvature of these 3D motion trajectories to represent actions which will capture the change in speed as well as direction. 

\begin{figure}
     \centering
     \subfloat[][]{\includegraphics[width=0.3\linewidth]{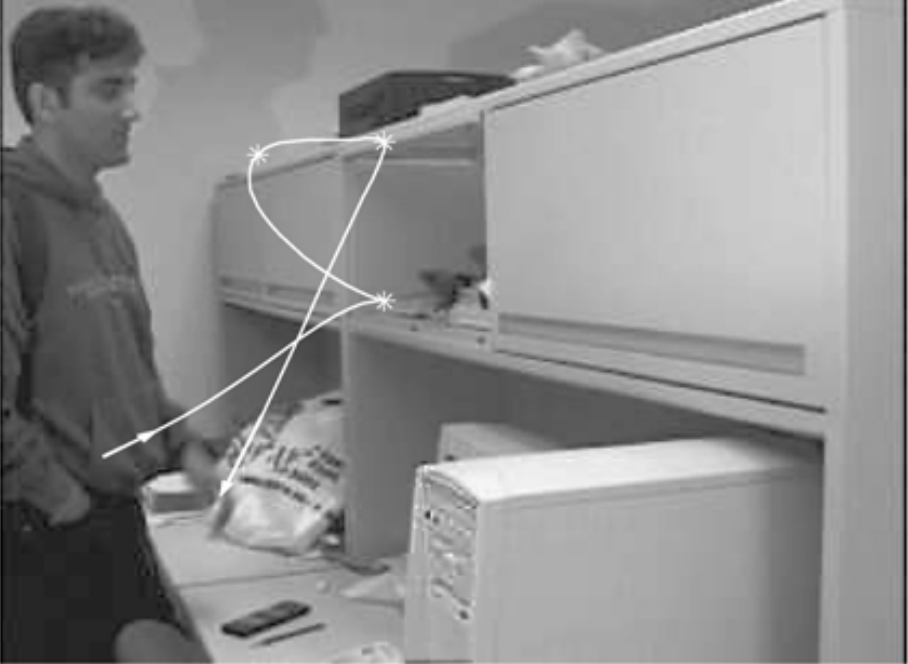} \label{fig_stc_sample_1}}
     \subfloat[][]{\includegraphics[width=0.3\linewidth]{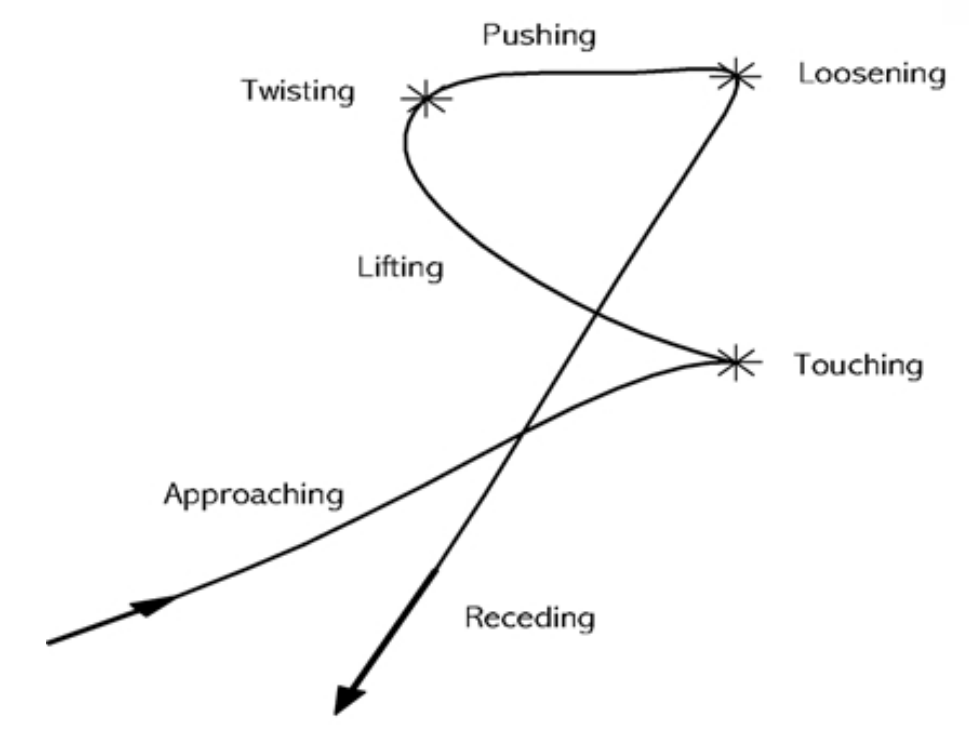} \label{fig_stc_sample_2}}
     \subfloat[][]{\includegraphics[width=0.3\linewidth]{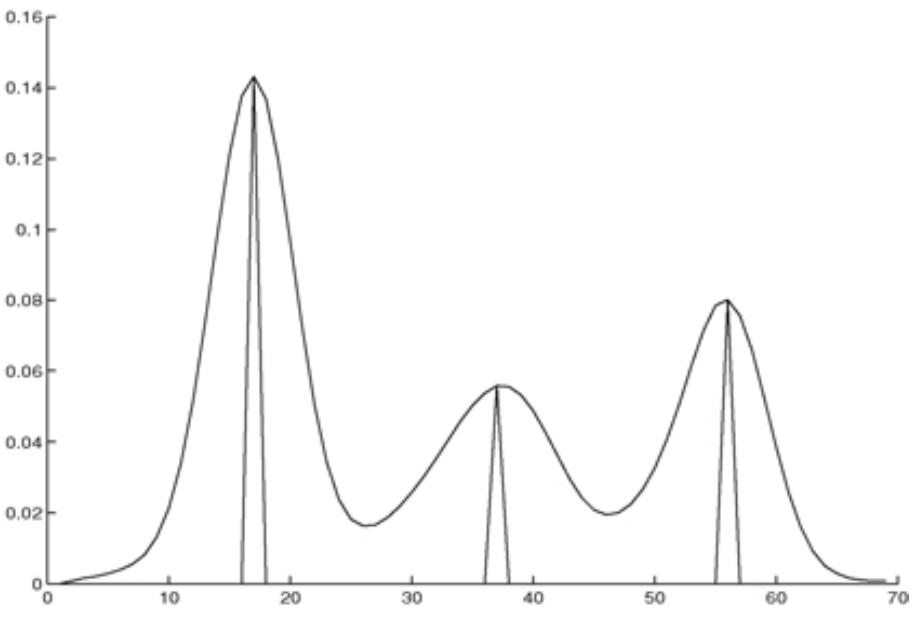} \label{fig_stc_sample_3}}
     \caption{(a) A sample video frame showing `opening a cabinet' action. A hand trajectory in white is superimposed on the image, (b) A representation of the trajectory in terms of instants and intervals, and (c) corresponding spatio-temporal curvature values and detected maximums (dynamic instants) \cite{rao2002view}.}
     \label{fig_stc_sample}
\end{figure}

A spatio-temporal curvature can be represented using instants which segment the motion trajectory into intervals. Instances indicate a significant change in the speed and direction in the motion trajectory and define motion boundaries \cite{rao2002view}. A special class of motion boundary, which is independent of starts and stops, is the dynamic instant that happens while performing an action. A dynamic instant represents a significant change in the motion characteristics and occurs only for one frame. It provides motion boundaries, called intervals, which represent the time period between two dynamic instants when the motion characteristics are not changing. In Figure \ref{fig_stc_sample_1}, we can observe a sample video frame for activity `opening a cabinet'. If we track the hand motion of the person while performing this action, we will get a spatio-temporal curvature shown in Figure \ref{fig_stc_sample_2}. It shows the instants as well as corresponding intervals in the motion trajectory. Figure \ref{fig_stc_sample_3} shows the spatio-temporal curvature values along with the detected dynamic instants in the motion trajectory.

\textbf{View-invariance:}
The discontinuities in 3D motion trajectory, which we perceive as instants, are always projected as discontinuities when projected in 2D curvatures \cite{rao2002view}. These instants which are maxima in spatio-temporal curvature will be view-invariant for most scenarios. Therefore, the number of instants in a spatio-temporal curvature is an important characteristic which will be view-invariant. The only exception will be the cases when there is a perfect alignment of the viewing direction with the plane where the action is being performed. In these cases, the position of the trajectory in consecutive frames will be projected to the same location in the frame resulting in a 2D trajectory, which is essentially a single point. Another important characteristic of instants in spatio-temporal curvature is the change in direction. This is known as sign characteristic, and it defines the direction of turns in action. It is very useful in distinguishing different actions captured from varying viewpoints. A clock-wise turn is represented by a `+' sign and counter clock-wise turn by `-' sign. 

These two characteristics of instants, number of instants and sign of instants, in a spatio-temporal curvature provide a view-invariant representation of actions. This representation can be used to determine whether two spatio-temporal curvature belongs to the same action class. The first requisite is a match between the number of instants and the sign sequence. Thereafter, we can compute a view-invariant similarity measure between the two trajectories. This can be performed using affine epipolar geometry, which will ensure viewpoint invariance in the similarity measure \cite{rao2002view}. However, there are some limitations with this approach. It requires an exact match between corresponding instants in the spatio-temporal curvature, which can be difficult as there can be false or missed instant detection. Also, this approach does not take into account the temporal information between the instants. These issues can be addressed using a view-invariant dynamic time warping to measure the similarity between two spatio-temporal curvatures \cite{rao2003view}. The view-invariant time-warping not only suppress the instant outliers, it also compensates for the variation in the execution-style \cite{rao2003invariance}. It can shrink the slow-motion trajectories, which are longer in temporal axis, and expand the fast motion trajectories which are relatively shorter.








\textbf{Limitations:}
The idea of spatio-temporal curvature is simple yet effective in extracting view-invariant representation for actions. However, there are certain limitations in this approach. The assumption that an action can be represented by a trajectory of single point in a video frame will not hold true for actions where full-body is involved. The skeleton joints in a human body will have a different motion for the same performed action. Therefore, this approach is limited to actions where the action can be approximated to a trajectory of single point at any time frame during the motion.

\subsection{Tracking joints}
A single point-based motion tracking of human actions can not be generalized for actions where multiple body joints are involved. It only carries motion information ignoring any shape or relative spatial information. Therefore, it is important to consider multiple human joints involved in the action while tracking the motion. The movement of different joints while performing an action is not independent of each other. The human body has certain anthropometric proportion, and there exist some geometric constraints between multiple anatomical landmarks such as body joints. This allows us to analyze human actions performed by different people using the semantic correspondence between human bodies.

\begin{figure}
     \centering
     \subfloat{\includegraphics[width=0.25\linewidth]{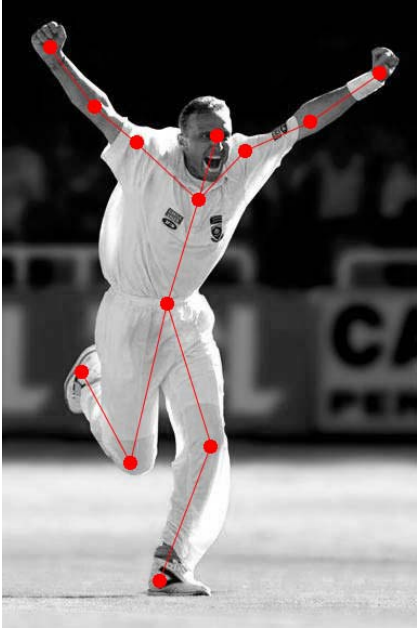} 
     } \quad \quad \quad \quad
     \subfloat{\includegraphics[width=0.25\linewidth]{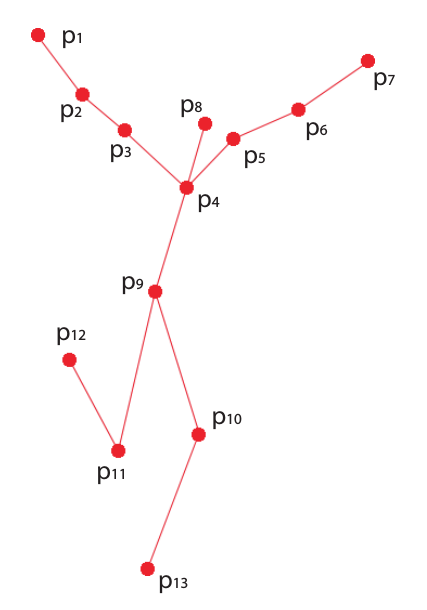} 
     }
     \caption{A sample video frame showing a pose and corresponding point-based pose representation \cite{gritai2004use}.}
     \label{fig_joints_sample1}
\end{figure}

The pose $\hat{X}$ of an actor while performing an action can be represented as a set of points in any given frame of a video. Here the pose is defined as $\hat{X} = \{X_1, X_2...X_n\}$, where $X_i = (x_i, y_i, z_i, \Lambda)^T$ are homogenous coordinates and there are $n$ such landmarks. Each point represents the spatial coordinate of an anatomical landmark on the human body, as shown in Figure \ref{fig_joints_sample1}. The body size and proportion vary greatly between different people and age groups. However, even though the human dimensional variability is substantial, it is not arbitrary \cite{gritai2004use}. Therefore, geometric constraints can be used between the pose of two different actors performing the same action. The proportion between two sets of points describing the pose of two different actors can be captured by a projective transformation. Suppose we have a set of points $\hat{X}$ and $\hat{Y}$ describing actors $A_1$ and $A_2$. Then the relationship between these two sets can be described by a matrix $M$ such that $X_i = MY_i$, where $i=1,2...n$ and $M$ is a 4x4 non-singular matrix. The transformation simultaneously captures the different pose of each actor as well as the difference in size/proportions of the two actors. There are two types of constraints which are useful for action recognition, postural constraints and action constraint \cite{gritai2004use}. The postures of two actors performing the same action at any time instant should be the same. This constraint allows us to recognize the action at each time instant by measuring the similarity between the postures. Along with this frame-wise constraint, another global constraint can be used on the point sets describing two actors if they are performing the same action. 

The anthropometric constraints in a human body allow the transformation of pose between two actors. Moreover, utilizing the postural and action constraints can help in recognizing action by measuring the similarity between two sets of points. The transformation and geometric constraints will address the issue of view-invariance in actions. However, this approach assumes the temporal alignment of poses for each frame in the video. This can be problematic as different actors will have a unique style of executing an action. Therefore, a temporal alignment is also required, which is invariant to temporal transformations. This can be done using dynamic time warping, which is particularly suited to action recognition as it is expected that different actors may perform portions of actions at a varying rate \cite{gritai2004use}.

\begin{figure}
     \centering
     \subfloat[][]{\includegraphics[width=0.45\linewidth]{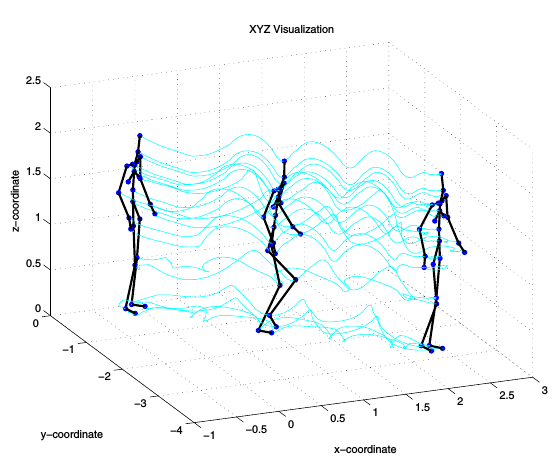} \label{fig_sample_joints_23}}
     \subfloat[][]{\includegraphics[width=0.45\linewidth]{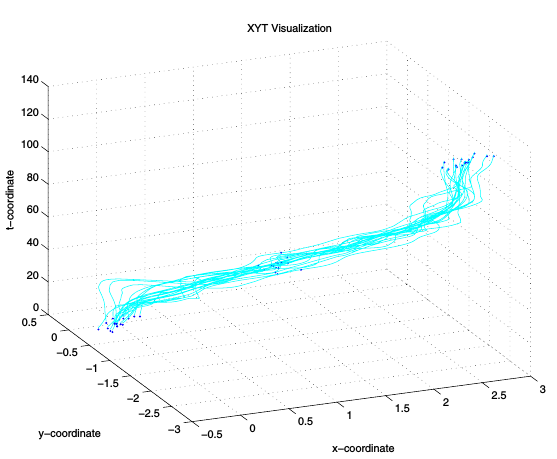} \label{fig_sample_joints_24}}
     \caption{Representation of an action in xyzt 4D space. (a) Action in xyz space, (b) action in xyt space \cite{sheikh2005exploring}.}    \label{fig_joints_sample2}
\end{figure}

The frame-wise representation of action decouples pose from its motion along the temporal domain. This approach can be effective to some level, but it ignores the temporal information focusing only on the order of poses, which limits its potential. We can model an action as a spatio-temporal construct since it is a function of time as well. It can be be represented as a set of points, $\hat{A} = \{X_1, X_2...X_p\}$, where $X_i = (x_i, y_i, z_i, t_i)^T$ are spatio-temporal coordinates, $p = mn$, where we have $m$ landmarks and there are $n$ recorded postures for that action \cite{sheikh2005exploring}. A sample action `walking' is shown in Figure \ref{fig_joints_sample2} in both xyz and xyt space at different time steps.

The variability associated with the execution of an action can be closely approximated by a linear combination of action basis in joint spatio-temporal space. An instance of an action can be defined as a linear combination of a set of action-basis $A_1, A_2...A_k$. Therefore, any instance of an action can be expressed as, $A^\prime = \Sigma_{i=1}^ka_iA_i$, where $a_i \in \mathbb{R}$ is the coefficient associated with the action-basis $A_i \in \mathbb{R}^{4xp}$. The space of an action, $A$ is the span of all its action basis. The variance captured by action basis can include different execution rates of the same action, different individual styles of performances as well as the anthropometric transformations of different actions. These action basis can be used to project a 4D pose point to its image (xyt) using a space-time projection matrix \cite{sheikh2005exploring}. These projections are useful in forming an action representation which can further be used for recognition of new instances. 

All the joints in the pose and all the time steps in a trajectory may not be required to identify the action category. We can select the joints dynamically based on the action and also choose fewer poses along the temporal domain known as canonical poses \cite{parameswaran2006view}. These empirically selected joints and canonical poses provide view-invariant trajectories which we call Invariance Space Trajectories (IST). These trajectories are also invariant representations of human actions and useful for action recognition.



\subsection{Spatio-temporal volume}
\begin{figure}
     \centering
     \subfloat[][Falling]{\includegraphics[width=0.24\linewidth]{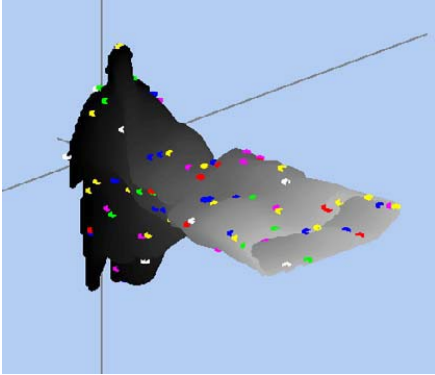} 
     }
     \subfloat[][Tennis stroke]{\includegraphics[width=0.24\linewidth]{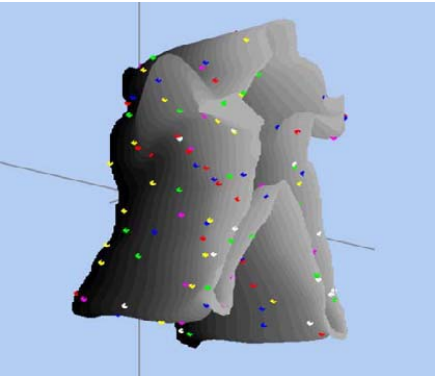} 
     }
     \subfloat[][Walking]{\includegraphics[width=0.24\linewidth]{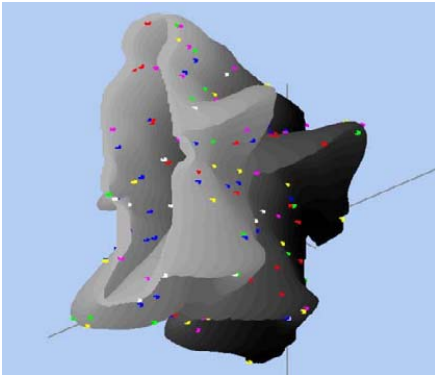} 
     }
     \subfloat[][Dancing]{\includegraphics[width=0.24\linewidth]{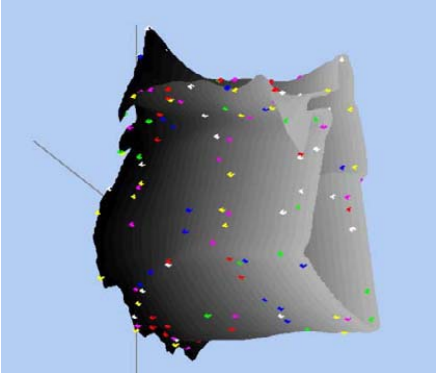} 
     }
     \caption{Generated Spatio-Temporal Volumes (STVs) for some sample actions. The color coded action descriptors are also shown corresponding to ridges (yellow), saddle ridges (white), peak (red), valley (pink), and saddle valleys (green) \cite{yilmaz2005actions}.}
     \label{fig_joints_sample3}
\end{figure}

The tracking of joints in the motion trajectories can be effective for action representation. However, as we discussed earlier, not all joints may be useful in recognizing the performed action. Another alternative to joints is the contour of the actor, which also captures the complete shape information. Tracking of actor contours will consider both shape and motion of the actor performing the action. When an actor performs some action in a 3D environment, the points on the outer boundary of the actor are projected as 2D (x,y) contour in the image plane. A sequence of 2D contours from frames with respect to time generates a Spatio-Temporal Volume (STV) in (x, y, t), which is a 3D object in the (x, y, t) space \cite{yilmaz2005actions}. The object contours in two consecutive frames can be tracked by finding a point correspondence between them. This can be done using a graph-theoretical approach by maximizing the match of a weighted bipartite graph. In Figure \ref{fig_joints_sample3}, we can observe some samples of generated spatio-temporal volumes for various actions. 

An STV can be considered as a manifold, and it will be nearly flat for small scales defined by a small neighborhood around a point. Therefore, it can be represented as a continuous action volume $\textbf{B}$ by computing plane equations in the neighborhood around a point. We define it using a 2D parametric representation by considering the arc length of the contour ($s$), which encodes the object shape, and time ($t$) which encodes the motion. The action volume is defined as $\textbf{B} = f(s,t) = [x(s,t), y(s,t), t]$. The parameters $s$ and $t$ can be used to generate trajectories and contours for any point in the object boundary. This representation is important for computing action descriptors corresponding to changes in direction, speed, and shape of parts of contours \cite{yilmaz2005actions}. This can be done by analyzing the surface type of a point in STV using Gaussian curvature and mean curvature. These surface types are important action descriptors, and for any given action they are called \textit{action sketch}. In figure \ref{fig_joints_sample3}, several of these action descriptors are superimposed on the STVs for various actions. The underlying curves of the contour and point trajectory for each action descriptor in the action sketch will have a maxima or a minima. It can be proved that the maxima/minima of the contour and the trajectory will not change by changing the viewpoint of the camera. Therefore, this representation is also invariant to any change in viewpoint.

\begin{figure}[h!]
\begin{center}
   \includegraphics[width=0.7\linewidth]{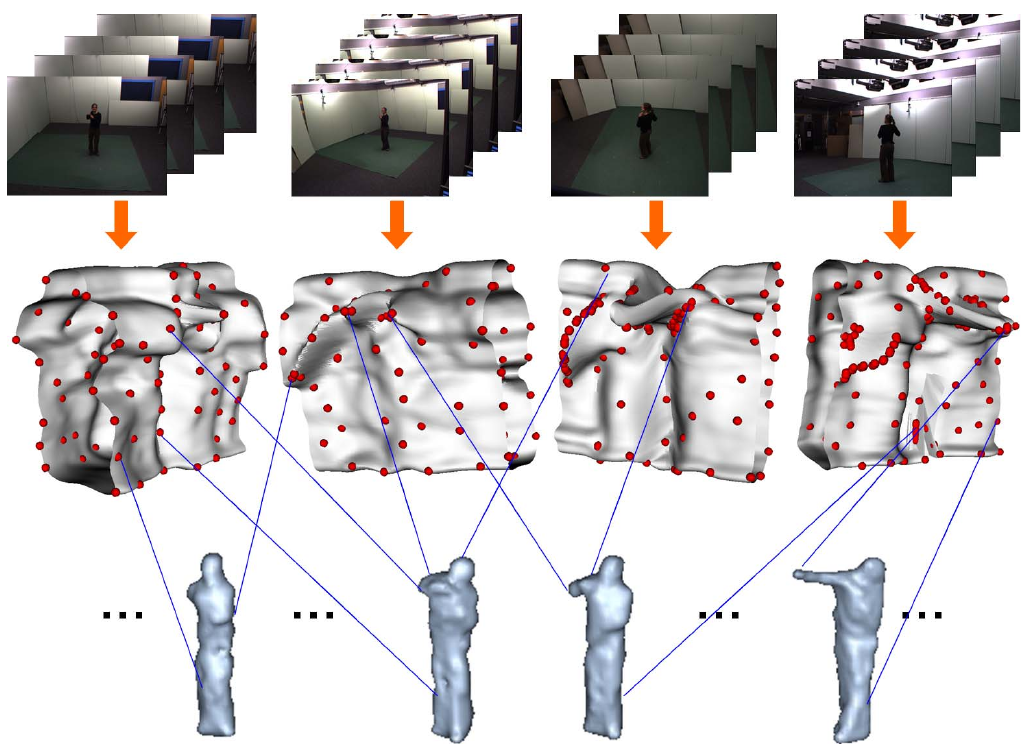}
\end{center}
\caption{Illustration of constructing 4D-AFM. The first row shows videos from four different views. The second row shows the STVs extracted from the videos and the locations of the spatio-temporal action features on the surface of STVs. These action features are mapped to the 4D action shape as shown in the third row \cite{yan2008learning}.}
\label{action_4d}
\end{figure}

The 3D volume of STV, along with the action descriptor derived from this volume, can be used for action recognition. The relation between two STVs is defined as  $x\mathcal{F}x^\prime = 0$, where $x$ and $x\prime$ are points on two different action sketches and $\mathcal{F}$ is a 3x3 fundamental matrix defining this relation. This relation estimates whether two different action sketches belong to the same action class and is a useful property for action recognition. This 3D volume can further be utilized to build a 4D action feature model(4D-AFM) \cite{yan2008learning}. This model elegantly encodes the shape and motion of actors observed from multiple views. A sample example for this model is shown in Figure \ref{action_4d}. This model enhances the view-invariant robustness of this approach, as features from multiple views are mapped to a unified model. Action recognition can be performed with this model based on the scores of matching action features from the action videos to the model points by exploiting pairwise interactions of features.



\subsection{Learning based methods}
The classical approaches focused on finding good features and utilize simple matching based techniques for action recognition. In learning based methods, both feature extraction and action recognition are performed jointly. The availability of multi-view activity datasets, such as IXMAS \cite{weinland2006free}, UWA3D MultiviewActivity II \cite{rahmani2016histogram}, Northwestern-UCLA \cite{wang2014cross} and NTU-RGBD \cite{shahroudy2016ntu}, has enabled the development of learning based robust methods for view-invariant feature learning. These features can be learned from different types of modalities, such as RGB \cite{liu2011cross,vyas2018time}, skeleton \cite{shahroudy2016ntu,rahmani2018learning}, depth \cite{li2018unsupervised,vyas2018time}, and optical flow \cite{li2018unsupervised}. 



We have seen earlier how motion trajectory is effective for extracting view-invariant descriptions. Apart from point-based and joint based trajectories, we can also extract dense trajectories from any action sequence. These trajectories provide dense points from each frame and track them using displacement information from a dense optical flow field. Given a dense trajectory of length $L$, a sequence $S$ is formed to represent the displacement vectors $\Delta P_t = (P_{t+1} - P_{t}) = (x_{t+1} - x_{t}, y_{t+1} - y_{t})$. The sequence is defined as a series of displacement vectors $(\Delta P_t,...,\Delta P_{t+L-1})$ which is normalized by $\sum_{i=t}^{t+L-1} ||{\Delta P_i}||$. Any action sequence can be represented using this dense motion trajectory description. 

These representations can be used to learn action-basis if we have a limited amount of samples. We discussed earlier how these action-basis can be useful for view-invariant action recognition. However, if we have a large amount of samples, computing action basis will be computationally expensive. An alternative is to perform clustering of these action descriptions and use the cluster centers as action representatives \cite{rahmani2018learning}. These cluster centers can further be used for describing new action trajectories using bag-of-word approach. If we have a sufficient number of samples from multiple views to compute these cluster centers, then we can train a fully connected neural network to perform non-linear transformation and learn view-invariant feature representation. These learned view-invariant features can then be used to perform robust action classification on novel and unseen views.


\textbf{Multi-task learning:}
The dense trajectories can be very effective for learning view-invariant representations for action recognition. However, they require an additional computation time, which is not suitable for systems with low latency. Convolutional neural network provides an efficient way to learn meaningful representations directly using the input video. These networks can be very effective in performing action classification on videos with previously seen views. However, they fail to generalize well for unseen views which are not present in the training data. This is caused due to the absence of view-invariance in the learned representation. This can be addressed by a multi-tasking approach where the representation is also utilized for some other task which enforces the network to learn a view-invariant representation. This task could be a prediction of optical flow from unseen view \cite{li2018unsupervised} or cross-view video synthesis \cite{vyas2018time}. Both these approaches enable the network to learn view-invariant feature representations and perform well on action recognition for unseen views.











\section{Datasets and Experimental Results}
The availability of public datasets for multi-view action recognition enables the research community to benchmark the research progress. There are four main public datasets which are widely used and provide action sequences captured from multiple viewpoints.  



\textbf{IXMAS:} The Inria Xmas Motion Acquisition Sequences (IXMAS) dataset \cite{weinland2006free} have videos for 11 action classes with actions performed 3 times by 10 actors. All actions have been captured using five camera views. 

\textbf{UWA3D MultiviewActivity II:} This dataset \cite{rahmani2016histogram} contains RGB videos of 30 human activities performed by 10 subjects with different scales. Each subject performed all the actions 4 times. Depth and Skeleton data captured using Kinect is also available with this dataset.

\textbf{Northwestern-UCLA:} Northwestern-UCLA Multiview 3D event dataset \cite{wang2014cross} contains RGB, depth and human skeleton data captured simultaneously by three Kinect cameras. This dataset has 10 action categories, and each action is performed by 10 actors. There are a total of 1493 action sequences. There are two main data splits: cross-subject (CS) and cross-view (CV) as suggested by \cite{wang2014cross}. View-invariant action recognition is examined with CV split where videos from the first two views are used for training, and the third view is for testing.

\textbf{NTU RGB-D:} This is the largest dataset for view-invariant human action recognition \cite{shahroudy2016ntu}. Along with RGB videos, depth and skeleton data is also available. It contains more than 56K videos and 4 million frames with 60 different action classes. There are a total of 40 different actors, who perform actions captured from 80 different viewpoints. 

Northwestern-UCLA and NTU RGB-D datasets provide multiple modalities, such as RGB, depth, and skeleton, along with multiple views to perform view-invariant action recognition. The performance of a method is evaluated by computing an accuracy score based on whether it predicts the correct action class or not. In Table \ref{table_ucla} and Table \ref{table2}, we can observe the performance of different methods using different modalities on Northwestern-UCLA and NTU RGB-D datasets. The cross-subject (CS) evaluation measure the performance on unseen subjects and the cross-view (CV) evaluation measure the performance on unseen views. The cross-subject performance is better than cross-view performance for most of the methods. This is mainly due to high variation in actions when seen from different viewpoints as compared with an unseen person where the appearance will have more variation. Therefore, the methods which focus on the motion will perform better than those which focus more on the appearance of the actor. 


\begin{table}[h!]
\centering
\begin{center}
\begin{tabular}{l|l|l|l}
\hline
Method & Modality & CS & CV \\
\hline\hline
%
Vyas et al. \cite{vyas2018time} & RGB & 87.5 & 73.2 \\ 
\hline
MST-AOG \cite{wang2014cross} & Depth & - & 53.6 \\
HOPC \cite{rahmani2016histogram} & Depth & - & 71.9 \\
CNN-BiLSTM \cite{li2018unsupervised} & Depth & - & 62.5 \\
\hline
R-NKTM \cite{rahmani2018learning} & Skeleton & - & 78.1 \\
\hline
MST-AOG \cite{wang2014cross} & RGB-S & 81.6 & 73.3 \\
\hline
\end{tabular}
\end{center}
\caption{{A comparison of cross-subject (CS) and cross-view (CV) action recognition on N-UCLA MultiviewAction3D dataset.}}
\label{table_ucla}
\end{table}

\begin{table}[h!]
\centering
\begin{center}
\begin{tabular}{l|l|l|l}
\hline
Method & Modality & CS & CV \\
\hline \hline
CNN-BiLSTM \cite{li2018unsupervised} & RGB & 55.5 & 49.3 \\
Vyas et al. \cite{vyas2018time} & RGB & 88.9 & 86.3 \\
\hline
CNN-BiLSTM \cite{li2018unsupervised} & Depth & 68.1 & 63.9 \\
Vyas et al. \cite{vyas2018time} & Depth & 79.4 & 78.7 \\
\hline
Shahroudy et al.\cite{shahroudy2016ntu} & Skeleton & 62.9 & 70.3 \\
\hline
CNN-BiLSTM \cite{li2018unsupervised} & Flow & 80.9 & 83.4 \\
\hline
DSSCA - SSLM\cite{shahroudy2018deep} & RGB-DS & 74.9 & - \\
\hline
\end{tabular}
\end{center}
\caption{{A comparison of cross-subject (CS) and cross-view (CV) action recognition on NTU-RGB+D dataset.
}}
\label{table2}
\end{table}


\section{Conclusion and open problems}
Action recognition for unknown and unseen views is a challenging problem. The tracking of motion trajectories have found to be successful so far but they are computationally expensive to extract and therefore not scalable for large-scale scenarios. The availability of large-scale multi-view datasets has enabled us to develop networks which are effective in recognition performance and can be trained end-to-end. However, their performance is limited by the variation of viewpoints in the available datasets. These datasets are lab curated, and actions are performed in a controlled environment. Apart from this, they are also limited in terms of variation in the number of available viewpoints. Therefore, it will be challenging to generalize the methods trained on these datasets to real-world scenarios. 


\nocite{*}
\bibliographystyle{plain}
\bibliography{template}

\begin{thebibliography}{10}

\bibitem{duarte2018videocapsulenet}
Kevin Duarte, Yogesh Rawat, and Mubarak Shah.
\newblock Videocapsulenet: A simplified network for action detection.
\newblock In {\em Advances in Neural Information Processing Systems}, pages
  7610--7619, 2018.

\bibitem{gritai2004use}
Alexei Gritai, Yaser Sheikh, and Mubarak Shah.
\newblock On the use of anthropometry in the invariant analysis of human
  actions.
\newblock In {\em Proceedings of the 17th International Conference on Pattern
  Recognition, 2004. ICPR 2004.}, volume~2, pages 923--926. IEEE, 2004.

\bibitem{li2018unsupervised}
Junnan Li, Yongkang Wong, Qi~Zhao, and Mohan Kankanhalli.
\newblock Unsupervised learning of view-invariant action representations.
\newblock In {\em Advances in Neural Information Processing Systems}, pages
  1254--1264, 2018.

\bibitem{liu2011cross}
Jingen Liu, M~Shah, B~Kuipers, and S~Savarese.
\newblock Cross-view action recognition via view knowledge transfer.
\newblock In {\em Proceedings of the 2011 IEEE Conference on Computer Vision
  and Pattern Recognition}, pages 3209--3216. IEEE Computer Society, 2011.

\bibitem{liu2017enhanced}
Mengyuan Liu, Hong Liu, and Chen Chen.
\newblock Enhanced skeleton visualization for view invariant human action
  recognition.
\newblock {\em Pattern Recognition}, 68:346--362, 2017.

\bibitem{parameswaran2006view}
Vasu Parameswaran and Rama Chellappa.
\newblock View invariance for human action recognition.
\newblock {\em International Journal of Computer Vision}, 66(1):83--101, 2006.

\bibitem{rahmani2016histogram}
Hossein Rahmani, Arif Mahmood, Du~Huynh, and Ajmal Mian.
\newblock Histogram of oriented principal components for cross-view action
  recognition.
\newblock {\em IEEE transactions on pattern analysis and machine intelligence},
  38(12):2430--2443, 2016.

\bibitem{rahmani2018learning}
Hossein Rahmani, Ajmal Mian, and Mubarak Shah.
\newblock Learning a deep model for human action recognition from novel
  viewpoints.
\newblock {\em IEEE transactions on pattern analysis and machine intelligence},
  40(3):667--681, 2018.

\bibitem{rao2003view}
C~Rao, A~Gritai, M~Shah, and T~Syeda-Mahmood.
\newblock View-invariant alignment and matching of video sequences.
\newblock In {\em Proceedings Ninth IEEE International Conference on Computer
  Vision}, 2003.

\bibitem{rao2003invariance}
Cen Rao, Mubarak Shah, and Tanveer Syeda-Mahmood.
\newblock Invariance in motion analysis of videos.
\newblock In {\em Proceedings of the eleventh ACM International Conference on
  Multimedia}, pages 518--527. ACM, 2003.

\bibitem{rao2002view}
Cen Rao, Alper Yilmaz, and Mubarak Shah.
\newblock View-invariant representation and recognition of actions.
\newblock {\em International Journal of Computer Vision}, 50(2):203--226, 2002.

\bibitem{shahroudy2016ntu}
Amir Shahroudy, Jun Liu, Tian-Tsong Ng, and Gang Wang.
\newblock Ntu rgb+ d: A large scale dataset for 3d human activity analysis.
\newblock In {\em Proceedings of the IEEE conference on computer vision and
  pattern recognition}, pages 1010--1019, 2016.

\bibitem{shahroudy2018deep}
Amir Shahroudy, Tian-Tsong Ng, Yihong Gong, and Gang Wang.
\newblock Deep multimodal feature analysis for action recognition in rgb+ d
  videos.
\newblock {\em IEEE Transactions on Pattern Analysis and Machine Intelligence},
  40(5):1045--1058, 2018.

\bibitem{sheikh2005exploring}
Yaser Sheikh, Mumtaz Sheikh, and Mubarak Shah.
\newblock Exploring the space of a human action.
\newblock In {\em Tenth IEEE International Conference on Computer Vision
  (ICCV'05) Volume 1}, volume~1, pages 144--149. IEEE, 2005.

\bibitem{vyas2018time}
Shruti Vyas, Yogesh~S Rawat, and Mubarak Shah.
\newblock Time-aware and view-aware video rendering for unsupervised
  representation learning.
\newblock {\em arXiv preprint arXiv:1811.10699}, 2018.

\bibitem{wang2014cross}
Jiang Wang, Xiaohan Nie, Yin Xia, Ying Wu, and Song-Chun Zhu.
\newblock Cross-view action modeling, learning and recognition.
\newblock In {\em Proceedings of the IEEE Conference on Computer Vision and
  Pattern Recognition}, pages 2649--2656, 2014.

\bibitem{weinland2006free}
Daniel Weinland, Remi Ronfard, and Edmond Boyer.
\newblock Free viewpoint action recognition using motion history volumes.
\newblock {\em Computer vision and image understanding}, 104(2-3):249--257,
  2006.

\bibitem{yan2008learning}
Pingkun Yan, Saad~M Khan, and Mubarak Shah.
\newblock Learning 4d action feature models for arbitrary view action
  recognition.
\newblock In {\em 2008 IEEE Conference on Computer Vision and Pattern
  Recognition}, pages 1--7. IEEE, 2008.

\bibitem{yilmaz2005actions}
Alper Yilmaz and Mubarak Shah.
\newblock Actions sketch: A novel action representation.
\newblock In {\em Proceedings of the 2005 IEEE Computer Society Conference on
  Computer Vision and Pattern Recognition (CVPR'05)-Volume 1-Volume 01}, pages
  984--989. IEEE Computer Society, 2005.

\bibitem{zheng2012cross}
Jingjing Zheng, Zhuolin Jiang, P~Jonathon Phillips, and Rama Chellappa.
\newblock Cross-view action recognition via a transferable dictionary pair.
\newblock In {\em bmvc}, volume~1, page~7, 2012.

\end{thebibliography}
\end{document}